\documentclass[letterpaper]{article} 
\usepackage{aaai2026}  
\usepackage{times}  
\usepackage{helvet}  
\usepackage{courier}  
\usepackage[hyphens]{url}  
\usepackage{graphicx} 
\usepackage{natbib}  
\usepackage{caption} 
\usepackage{algorithm}
\usepackage{algorithmic}

\usepackage{newfloat}
\usepackage{listings}
\DeclareCaptionStyle{ruled}{labelfont=normalfont,labelsep=colon,strut=off} 
\floatstyle{ruled}
\newfloat{listing}{tb}{lst}{}
\floatname{listing}{Listing}
\newcommand{\SystemName}{%
  \textbf{MAP}%
}

\usepackage{upgreek}

\usepackage{color, colortbl}
\definecolor{codegreen}{rgb}{0,0.6,0}
\definecolor{codegray}{rgb}{0.5,0.5,0.5}
\definecolor{codepurple}{rgb}{0.58,0,0.82}
\definecolor{backcolour}{RGB}{230, 230, 230}
\definecolor{yelloworange}{RGB}{242, 182, 70}\definecolor{yellowtext}{RGB}{249, 191, 0}
\definecolor{LightCyan}{rgb}{0.88,1,1}
\definecolor{LightRed}{RGB}{255, 204, 203}
\usepackage{amsmath}

 \usepackage{amsmath}
 \usepackage{amsfonts}
 \usepackage{pifont}
 \usepackage{amssymb}

\usepackage{multirow}
\usepackage{booktabs}

\usepackage[titletoc]{appendix}

\title{ Machine Pareidolia: Protecting Facial Image with Emotional Editing}
\author{
    Binh M. Le,
    Simon S. Woo\thanks{Simon S. Woo is the corresponding author.} 
}
\affiliations{
   Sungkyunkwan University, Suwon, South Korea\\
    bmle@g.skku.edu, swoo@g.skku.edu
}

\usepackage{bibentry}

\begin{document}

\maketitle

\begin{abstract}
   The proliferation of facial recognition (FR) systems has raised privacy concerns in the digital realm, as malicious uses of FR models pose a significant threat. Traditional countermeasures, such as makeup style transfer, have suffered from low transferability in black-box settings and limited applicability across various demographic groups, including males and individuals with darker skin tones. 
   To address these challenges, we introduce a novel facial privacy protection method, dubbed \textbf{MAP}, a pioneering approach that employs human emotion modifications to  disguise original identities as target identities in facial images. Our method uniquely fine-tunes a score network to learn dual objectives, target identity and human expression, which  are jointly optimized through gradient projection to ensure convergence at a shared local optimum. Additionally, we enhance the perceptual quality of protected images by applying local smoothness regularization and optimizing the score matching loss within our network. Empirical experiments demonstrate that our innovative approach surpasses previous baselines, including noise-based, makeup-based, and freeform attribute  methods, in both qualitative fidelity and quantitative metrics. Furthermore, \textbf{MAP} proves its effectiveness against an online  FR API  and shows advanced adaptability in uncommon photographic scenarios. 
 
\end{abstract}
\section{Introduction}
\label{sec:intro}
\begin{figure}[t]
\centering
\includegraphics[width=0.48\textwidth]{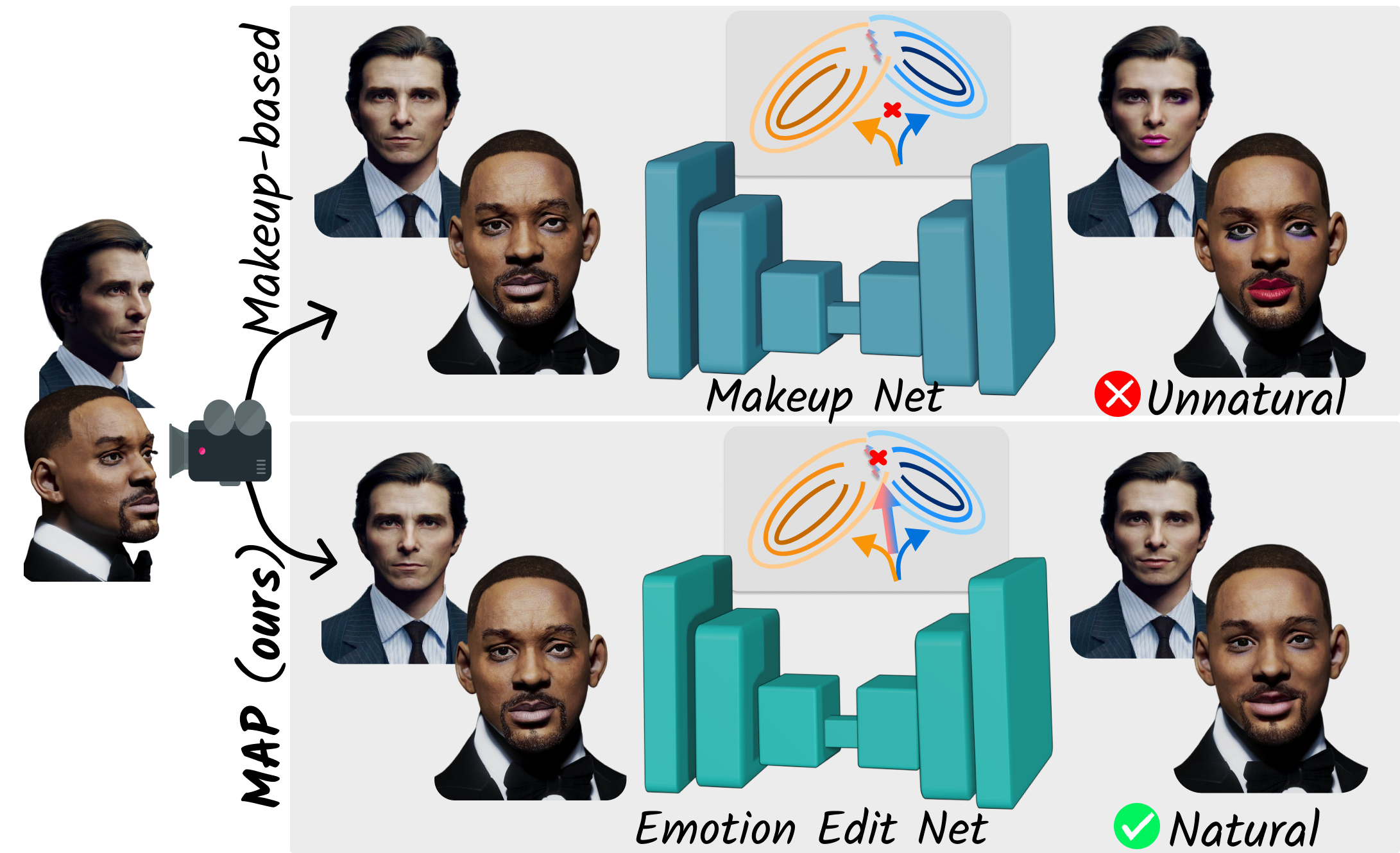}
\caption{{Comparison of makeup-based baselines with our \SystemName}. Baseline methods (top) use makeup styles obscure original identities, optimizing objectives independently, which reduces efficiency and limits applicability across demographics (\textit{e.g.}, males). In contrast, \SystemName\ (bottom) leverages human emotion modifications and a unified optimization strategy to disguise original identities as target identities, achieving universal robustness.
}
\label{fig:thumbnail}

\end{figure}
Recent advances in deep learning-based face recognition (FR) systems have enabled their widespread adoption in applications like biometrics \cite{meden2021privacy}, security \cite{wang2017face}, and criminal investigation \cite{phillips2018face}. However, these advancements also pose significant privacy risks in the digital realm. Malicious uses of FR, such as unauthorized surveillance \cite{wenger2023sok, hill2024facerecognition}, tracking relationships, and monitoring activities on social platforms \cite{hill2022secretive, shoshitaishvili2015portrait}, highlight the urgent need for effective methods to protect facial images from unauthorized FR systems.

An effective facial privacy protection method should achieve an optimal balance {between maintaining} \textit{natural appearance} and {ensuring} \textit{robust privacy}. Early approaches  \cite{zhou2024securely,  zhou2023advclip, zhong2022opom, tip_im} overlaid bounded adversarial perturbations on original images, but these often rendered images unnatural, negatively impacting user experience. Subsequent methods \cite{na2022unrestricted, kakizaki2019adversarial, amt_gan, clip2protect, diffam} employed unrestricted adversarial examples to thwart FR systems. Among these, adversarial makeup style transfer \cite{adv_makeup, amt_gan, clip2protect, diffam} has gained substantial attention for its natural edits. However, existing makeup-based techniques face key limitations (Fig. \ref{fig:thumbnail} - Top): (i) Unnatural edits: makeup style transfer yields unnatural results, particularly for demographics such as males and darker-skinned individuals; (ii) Suboptimal dual-task learning: simultaneously optimizing unrelated objectives (adversarial identity and makeup style) can trigger a tug-of-war between conflicting gradients, causing negative transfer and leading to unnatural and non-robust outcomes.

Recently, diffusion models \cite{ho2020denoising} have gained attention in image editing due to their training stability and ability to capture the full data distribution through their simple denoising process \cite{dhariwal2021diffusion}. Guided diffusion models, enhanced with pretrained CLIP text embeddings \cite{clip_model}, have also shown promising results across various applications \cite{kim2022diffusionclip, liu2023more, diffam}. Nevertheless, effectively injecting adversarial noise into attribute edits for facial privacy while minimally altering the original image to maintain utility remains an open challenge.

To this end, we propose MAchine Pareidolia (\SystemName) - a novel solution that exploits the human psychological marvel of \textit{pareidolia}, where we perceive familiar faces in ambiguous patterns~\cite{liu2014seeing}. \SystemName~ transforms facial action units to dupe FR systems into recognizing target identities, leveraging their inherent tendency to misread subtle emotional tweaks. Unlike prior approaches, our medium-to-high-frequency emotion {modifications} are seamless across all demographics, avoiding the unnatural, exaggerated appearance of makeup-based techniques and global mismatches like skin tone shifts (Fig. \ref{fig:thumbnail}). Inspired by research on multitasking learning \cite{hsieh2024careful, yu2020pcgrad}, we introduce a synergistic gradient adjustment strategy {in the following way} to resolve the conflict between adversarial identity and expression gradients: whenever layer-wise minibatch gradients  oppose empirical gradients of the other loss, we decompose the minibatch gradients into parallel and orthogonal components relative to the empirical gradients, retaining only the orthogonal component to forge a cohesive path to a shared optimal region. To preserve the natural allure of protected images, we use Laplacian smoothness regularization, delicately maintaining the relative positions of facial landmarks to prevent catastrophic distortions.

Extensive experiments on the CelebA-HQ and LADN datasets showcase \SystemName 's superior privacy protection - boosting black-box success rates by up to 11\% - while delivering exceptional perceptual quality and universal applicability across different demographics and photographic scenarios. Moreover, \SystemName{} exhibits a more favorable balance between perceptual quality and identity obfuscation when compared to  recent free-form approaches. Our contributions are summarized as follows:
\begin{itemize}
    \item We propose \SystemName, a psychology-inspired approach that subtly changes facial expressions to disguise original identities as target ones, thwarting malicious FR systems.
    \item We introduce a novel synergistic gradient adjustment strategy to harmonize identity and emotion edits, guiding the model to a shared optimal region. To prevent catastrophic distortions from expression edits, we propose Laplacian smoothness regularization to preserve the relative positions of facial landmarks.
    \item Extensive experiments on diverse datasets demonstrate our \SystemName{} outperforms prior work, including freeform protection baselines, showcasing \SystemName 's applicability across demographics and photographic styles.
\end{itemize}

\section{Related Works}
\label{sec:related_work}
\textbf{{Face Protection Method}.}Due to privacy concerns regarding user identities on online social platforms, various obfuscation methods have been proposed to conceal identities \cite{meden2021privacy}. Lately, the advent of deep neural networks has facilitated more advanced approaches to shield users from unauthorized facial recognition (FR) systems. Early strategies often employed noise-based adversarial samples \cite{zhou2024securely,  zhou2023advclip, zhong2022opom, tip_im, oh2017adversarial}, which involved adding carefully designed adversarial perturbations to the original face images to mislead hostile FR models. \citet{oh2017adversarial} developed a game-theoretical framework to derive guarantees on privacy levels in white-box settings. Additionally, TIP-IM \cite{tip_im} created adversarial identity masks that can be imposed on facial images to obscure the original identity against black-box FR models. However, such perturbations are usually noticeable to observers and can compromise the user experience.

Recently, strategies that utilize unbounded adversarial samples, which do not constrain the perturbation norm in pixel space, have emerged, resulting in improved  image quality \cite{adv_makeup, amt_gan, clip2protect, diffam, diffprotect, gift_mm}. Among these, makeup-based methods conceal perturbations under the guide of natural makeup features. However, this approach may not be suitable for certain demographic groups, such as males. Another line of research employs generative models to traverse various image attributes, thereby creating adversarial images \cite{joshi2019semantic, khedr2023semantic, diffprotect,diff_privacy,du2024multi}. Specifically, \citet{gift_mm} propose a two-step optimization in the low-dimensional latent space of a GAN model \cite{stylegan} to construct a protected image. Similarly, DiffProtect \cite{diffprotect} optimizes the  latent vector  of the source image adversarially to mask the original identity. However, since latent vectors encode a wide range of entangled semantic features, these methods can inadvertently alter global features such as image lighting or saturation, misaligning edited faces with the original scene.

\textbf{Diffusion Models and Multimodal Guidance}. 
The rise of probabilistic generative models, particularly score-based diffusion models \cite{ho2020denoising}, has been fueled by their training stability and scalability, sparking their adoption across diverse vision tasks. These include image generation \cite{rombach2022high, dhariwal2021diffusion}, image editing \cite{meng2021sdedit, wang2023stylediffusion},  image restoration \cite{xia2023diffir}, \textit{ect.}. To control generated content and style, multimodal approaches leverage pretrained text or vision encoders to guide the score network \cite{zhang2023adding, kim2022diffusionclip}. Notably, CLIP, a dual network pretrained on text-image pairs, is widely used for style transfer by navigating its shared latent space. However, effectively injecting adversarial noise into CLIP-guided style edits to enhance privacy remains an open challenge.
\section{Methods}
\subsection{Problem Statement}
Let $\boldsymbol{x} \in \mathcal{D}$ be an original face image. A pretrained face recognition model maps input image $\boldsymbol{x}$ to a hypersphere space as $z = f(\boldsymbol{x}) : \mathcal{D} \rightarrow \mathcal{Z} \subset \mathbb{R}^{d}$. The similarity between two facial images $\boldsymbol{x}_i$ and $\boldsymbol{x}_j$, measured by $f$, is denoted by $\Phi(z_i, z_j | f)$ and typically adopts cosine similarity between $z_i$ and $z_j$. Throughout the paper, we use \(\bar{v}\) to denote the Euclidean normalization (or unit vector) of a general vector \(v\).  Let \(\mathcal{L}\) and \(\mathcal{L}^{\mathcal{B}_t}\) represent the expected objective value over the entire dataset \(\mathcal{D}\) and a subset \(\mathcal{B}_t \subset \mathcal{D}\), respectively.

Black-box attacks on face recognition systems are generally categorized into targeted attacks (impersonation attacks) and non-targeted attacks (dodging attacks). Following the settings in \cite{diffam}, we specifically focus on targeted attacks for more efficient protection of identity images.  In this scenario, an operator \(\mathcal{T}\) parameterized by \(\boldsymbol{w}\) transforms an original face image \(\boldsymbol{x}_o\) into a perturbed image \(\boldsymbol{x}_p\), where \(\boldsymbol{x}_p = \mathcal{T}_{\boldsymbol{w}}(\boldsymbol{x}_o)\), such that \(\boldsymbol{x}_p\) successfully impersonates a target face \(\boldsymbol{x}_t\), without knowledge of the target face recognition model. At the same time, we ensure that the transformed image \(\boldsymbol{x}_p\) does not significantly deviate from the natural image manifold, maintaining its usability. Formally, the optimization problem that we aim to solve is: 
\begin{align}
        \max_{\boldsymbol{w}}&\text{ } \Phi(\underbrace{f({\mathcal{T}_{\boldsymbol{w}}(\boldsymbol{x}_o)})}_{z_p}, \underbrace{f(\boldsymbol{x}_t)}_{z_t})  \text{~\textit{s.t.}~} \mathcal{H}(\mathcal{T}_{\boldsymbol{w}}(\boldsymbol{x}_o), \boldsymbol{x}_o) \leq \varepsilon 
\end{align}
 where $\varepsilon$ bounds the extent of image modifications. For a noise-based approach, the \(\ell_p\) norm of \((\boldsymbol{x}_p - \boldsymbol{x}_o)\) is typically used for \(\mathcal{H}\), though this can introduce noticeable artifacts that compromise user experience. In the above objective equation, $\Phi(z_p, z_t | f)$ is generally unknown since $f$ is a black-box model, and we substitute  it with a set of $M$ surrogate models $\{f_i\}_{i=1}^{M}$ for optimization.

\subsection{MAP: Emotion-based Approach}
\begin{figure}[t]
\centering
\includegraphics[width=0.44\textwidth]{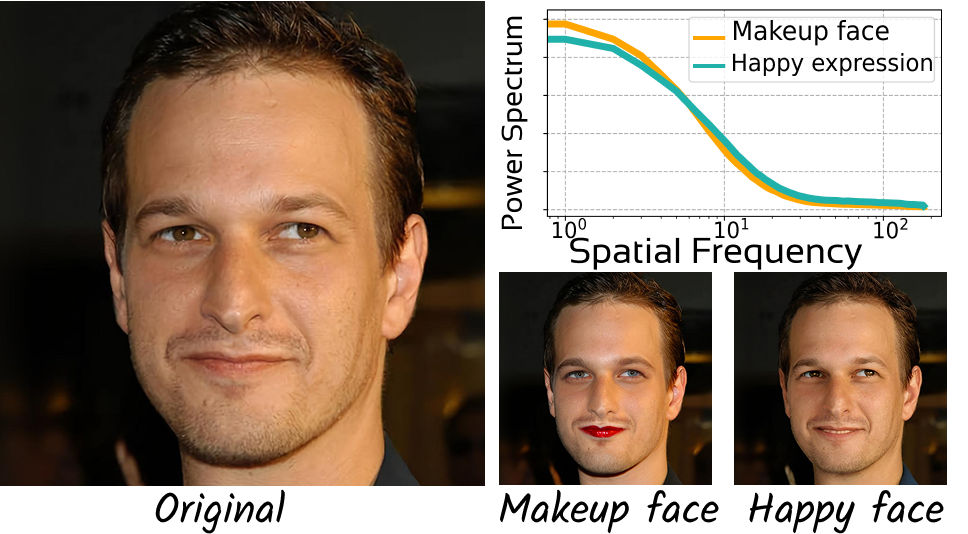}
\caption{{Comparison between makeup transfer-based and emotion-based approaches in terms of frequency changes} (with azimuthal integral). Makeup transfer primarily edits the image in the low-frequency range, while our emotion-based approach targets medium to high frequencies, resulting in a more natural appearance and being better suited for obfuscating identity adversarial noises.
}
\label{fig:frequency}
\end{figure}

In response to the limitations of prior works  as identified in previous sections, our proposed method leverages emotion-based editing that exposes \textbf{three solid advantages} compared with prior studies: This technique involves adjusting subtle facial action units that correspond to different emotional states without altering core demographic attributes, thus demographic-agnostic. Secondly, unlike makeup transfer, which primarily introduces low-frequency artifacts, action units involve medium to high frequencies, making them suitable for infusing subtle adversarial noises, as illustrated in Fig. \ref{fig:frequency}. Lastly, as a consequence of the second advantage, our method has unnoticeable interference in non-face areas, maintaining the faithfulness of the original image. Hence, our objective is to inject adversarial noises into the emotional change of a person and dually optimize them, while maintaining the original looks. 

Given a pretrained  score network \(\boldsymbol{s_w}\), let \(\boldsymbol{x}_o\) be a face image drawn from a training dataset \(\mathcal{D}\). We first transform \(\boldsymbol{x}_o\) into a stochastic latent variable of \(\boldsymbol{s_w}\) at timestep \(\tau\) using a Gaussian transition \cite{ho2020denoising}:
\begin{equation}
    \boldsymbol{x}^\tau_o = \sqrt{\alpha_\tau}\boldsymbol{x}_o + (1-\alpha_\tau)\boldsymbol{r}, \quad \boldsymbol{r}\sim\mathcal{N}(\boldsymbol{0}, \boldsymbol{I}),
\end{equation} 
where $\alpha_\tau=\Pi_{i=1}^{\tau}(1-\beta_i)$ with $\beta_i$'s are variance schedule. To generate an edited image, we adopt fast detemisnistic reverse DDIM process \cite{songdenoising}:
\begin{equation}
    \boldsymbol{x}^{\tau-1}_o = \sqrt{\alpha_{\tau-1}}\tilde{x_o} + \sqrt{1-\alpha_{\tau-1}} \boldsymbol{s_w}(\boldsymbol{x}^{\tau}_o, \tau),
\end{equation}
where $\tilde{x_o}=(\boldsymbol{x}^{\tau}_o-\sqrt{1-\alpha_{\tau-1}} \boldsymbol{s_w}(\boldsymbol{x}^{\tau}_o, \tau))/\sqrt{\alpha_t}$. After \(\tau\) backward steps, we obtain the edited image, denoted as the protected image \(\boldsymbol{x}_p = \boldsymbol{x}^0_o\).

\textbf{Dual Objectives}. During the training of \(\boldsymbol{s_w}\) on \(\mathcal{D}\), we employ a set of \(M\) surrogate face recognition models \(\{f_i\}_{i=1}^{M}\) to bring \(\boldsymbol{x}_p\) and \(\boldsymbol{x}_t\) closer together. To achieve this, we minimize the cosine discrepancy between \(\boldsymbol{x}_p\) and \(\boldsymbol{x}_t\) in the latent space of each \(f_i\) using an angular divergence loss:
\begin{align}
    \mathcal{L}_{\text{A}}(\boldsymbol{w}) = \mathbb{E}_{\boldsymbol{x}_o\sim\mathcal{D}}\left(\frac{1}{M}\sum_{i=1}^{M} 1-\langle \overline{f_i}(\boldsymbol{x}_p),  \overline{f_i}(\boldsymbol{x}_t) \rangle \right ), 
\label{eqn:adv}
\end{align}
 where $\langle \cdot, \cdot\rangle$ denotes the inner product, and $\overline{f_i}(\boldsymbol{x})$ is the $\ell_2$ normalization of $f_i(\boldsymbol{x})$. Meanwhile, to conceal visual adversarial artifacts stemming from Eq. \ref{eqn:adv} in $\boldsymbol{x}_p$, we apply a demographic-agnostic transformation through emotional alteration of $\boldsymbol{x}_p$. Unlike previous methods \cite{clip2protect, diffam}, this transformation subtly adjusts action units \cite{actionunits, tian2001recognizing}, targeting their medium-to-high-frequency components, to effectively mask adversarial artifacts while preserving the utility of $\boldsymbol{x}_p$. Inspired by \cite{stylegannada}, we leverage the semantic information encapsulated in the pretrained CLIP \cite{clip_model} model. This induces alignment between the embeddings of the reference image $\boldsymbol{x}_o$ and the generated image $\boldsymbol{x}_p$, and those of the reference text \texttt{"source"} and target text \texttt{"target"} within the CLIP space. Formally, the emotion objective is defined as follows:
\begin{align}
    \Delta_{txt}&= \mathcal{E}_{txt}(\texttt{"target"}) -  \mathcal{E}_{txt}(\texttt{"source"}), \nonumber \\
    \Delta_{vis}&= \mathcal{E}_{vis}({\boldsymbol{x}_p}) -  \mathcal{E}_{vis}({\boldsymbol{x}_o}), \nonumber \\
    \mathcal{L}_{\text{E}}(\boldsymbol{w}) &=  \mathbb{E}_{\boldsymbol{x}_o\sim\mathcal{D}} \left (1 - \langle \overline{\Delta}_{txt}, \overline{\Delta}_{vis} \rangle \right ), \label{eqn:emoji}
\end{align}
where $\mathcal{E}_{\text{txt}}$ and $\mathcal{E}_{\text{vis}}$ are textual and visual encoders, respectively. Here, we define \texttt{"target"} as \textit{"a photo of [emotion] face"} and \texttt{"source"} as \textit{"a photo of face"}.

\textbf{Efficient Dual Objective Optimization}.
Thus far, we have optimized two objectives, \(\mathcal{L}_{\text{A}}\) and \(\mathcal{L}_{\text{E}}\), simultaneously within the diffusion model \(\boldsymbol{s_w}\). However, since these objectives in Eqs. \ref{eqn:adv} and 
 \ref{eqn:emoji} address distinct transformations: \textit{identity} and \textit{emotion}, respectively, their concurrent optimization may lead to a conflict between opposing gradients. This tug-of-war may yield suboptimal results; for instance, while \(\boldsymbol{x}_p\) may effectively mask the original identity, the resulting appearance could appear unnatural, or vice versa. Inspired by recent studies \cite{hsieh2024careful, yu2020pcgrad}, we propose managing the gradient updates from both losses at every layer of the network. 
\begin{figure}[t]
\centering
\includegraphics[width=0.36\textwidth]{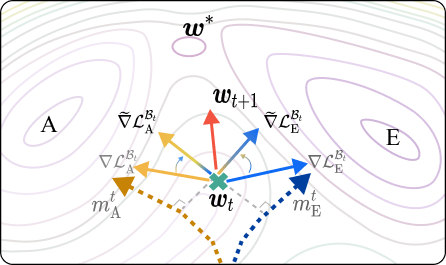}
\caption{\textbf{Illustration of our dual objectives optimization strategies.} Naively optimizing unrelated tasks (identity \textcolor{yelloworange}{$\boldsymbol{\leftarrow}$} and emotion \textcolor{blue}{$\boldsymbol{\rightarrow}$}) may lead to negative transfer, canceling out each other's gradients. Our approach can render a new update (\textcolor{red}{$\boldsymbol{\uparrow}$}) that helps guide the model towards optimal values.
}
\label{fig:grad_proj}
\end{figure}
Let \(\nabla\mathcal{L}_{\text{A}}(\boldsymbol{w}_l)\) and \(\nabla\mathcal{L}_{\text{E}}(\boldsymbol{w}_l)\) represent the gradients at the \(l^{th}\) layer of \(\boldsymbol{s_w}\), derived from the empirical losses in Eqs. \ref{eqn:adv}  and \ref{eqn:emoji}, respectively. At training iteration \(t\), with a minibatch \(\mathcal{B}_t\) of images, we compute gradients for each loss as \(\nabla\mathcal{L}^{\mathcal{B}_t}_{\text{A}}(\boldsymbol{w}_l)\) and \(\nabla\mathcal{L}^{\mathcal{B}_t}_{\text{E}}(\boldsymbol{w}_l)\). We then adjust any conflicting gradients, \(\nabla\mathcal{L}^{\mathcal{B}_t}_{\text{A/E}}(\boldsymbol{w}_l)\), which exacerbate the discrepancies between \(\nabla\mathcal{L}_{\text{A}}(\boldsymbol{w}_l)\) and \(\nabla\mathcal{L}_{\text{E}}(\boldsymbol{w}_l)\). Formally:
\begin{align}
    \widetilde{\nabla}\mathcal{L}^{\mathcal{B}_t}_{\text{A}} = \nabla \mathcal{L}^{\mathcal{B}_t}_{\text{A}} - \min \left \{ 0,  \langle\nabla\mathcal{L}^{\mathcal{B}_t}_{\text{A}},\overline{\nabla\mathcal{L}}_{\text{E}}\rangle  \right \} \cdot \overline{\nabla\mathcal{L}}_{\text{E}},
    \label{eqn:proj_a}\\
    \widetilde{\nabla}\mathcal{L}^{\mathcal{B}_t}_{\text{E}} = \nabla \mathcal{L}^{\mathcal{B}_t}_{\text{E}} - \min \left \{ 0,  \langle \nabla\mathcal{L}^{\mathcal{B}_t}_{\text{E}},\overline{\nabla\mathcal{L}}_{\text{A}}\rangle \right \} \cdot \overline{\nabla\mathcal{L}}_{\text{A}}.
    \label{eqn:proj_e}
\end{align}
  Intuitively, for $ \nabla\mathcal{L}^{\mathcal{B}_t}_{\text{A}}(\boldsymbol{w}_l)$, whenever it forms an angle with 
 $\nabla\mathcal{L}_{\text{E}}(\boldsymbol{w}_l)$ greater than $90^{\circ}$ (thus increasing update differences), we decompose $ \nabla\mathcal{L}^{\mathcal{B}_t}_{\text{A}}$ into two independent components: one parallel and one orthogonal with  $\nabla\mathcal{L}_{\text{E}}(\boldsymbol{w}_l)$. We retain only the orthogonal component as described by Eq. \ref{eqn:proj_a}. {For} $\nabla\mathcal{L}_{\text{A/E}}(\boldsymbol{w}_l)$, as computing the full gradient on the entire dataset is computationally prohibitive, we estimate these gradients using an {exponential} moving average (EMA) that accumulates historical minibatch gradients:
\begin{equation}
\boldsymbol{m}^t_{\text{A/E}}(\boldsymbol{w}_l) = \lambda\boldsymbol{m}^{t-1}_{\text{A/E}}(\boldsymbol{w}_l)+(1-\lambda)\nabla\mathcal{L}^{\mathcal{B}_t}_{\text{A/E}} (\boldsymbol{w}_l), 
    \label{eqn:ema}
\end{equation}
where $\lambda$ is a hyperparameter. To ensure Eq. \ref{eqn:ema} closely approzimates  $\nabla\mathcal{L}_{\text{A/E}}$, we present the following theorem.

\textbf{\textit{Momentum Bound Theorem.}} \textit{Suppose that the loss functions in Eqs. \ref{eqn:adv} and \ref{eqn:emoji} satisfy the following assumptions: (1) their gradients $\nabla \mathcal{L}_{\text{A/E}}(\boldsymbol{w})$ are bounded, i.e., $||\nabla \mathcal{L}_{\text{A/E}}(\boldsymbol{w}^t)||\leq G$; (2) the stochastic gradients are L-Lipschitz, i.e, $||\nabla \mathcal{L}_{\text{A/E}}(\boldsymbol{w}^t)-\nabla \mathcal{L}_{\text{A/E}}(\boldsymbol{v}^{ t})|| \leq L||\boldsymbol{w}^{t}-\boldsymbol{v}^{  t} ||$, $\forall \boldsymbol{w}^{t}, \boldsymbol{v}^{t}$; (3) the gradient's variance is bounded, i.e., there exists a constant $M>0$ for any data batch $\mathcal{B}_t$ such that $\mathbb{E}	\left[\| \nabla \mathcal{L}_{\text{A/E}}^{\mathcal{B}_t}(\boldsymbol{w}) - \nabla \mathcal{L}_{\text{A/E}}(\boldsymbol{w})  \|_2^2 \right] \le M, \quad \forall \boldsymbol{w} \in  \mathbb{R}^d.$
   Assume that $\boldsymbol{s_w}$  uses SGD as the base optimizer with a learning rate $\eta$ to update the model parameter with adjusted gradients in Eqs. \ref{eqn:proj_a} and \ref{eqn:proj_e}. Then, by setting $\lambda=1-\left( M^{1/2}/LG\right)^{2/3}\eta^{2/3}$, after $T > C^\prime \eta^{-2/3}$  training iterations ($C'$ is a constant), with probability $1-\delta$, we obtain:}   
    $\|\boldsymbol{m}^T_{A/E} - \nabla \mathcal{L}_{\text{A/E}}(\boldsymbol{w}^T)\|_2  \leq  \mathcal{O}(\eta^\frac{1}{3} L^\frac{1}{3} G^\frac{1}{3} M^\frac{1}{3} \log^\frac{1}{2}(1/\delta))$.

This theorem establishes that   the error bound between the full gradient \(\nabla \mathcal{L}(\boldsymbol{w}^t_\text{A/E})\) and its EMA estimate \(\boldsymbol{m}^t_\text{A/E}\), at \(\mathcal{O}(\gamma^{1/3})\), reduces to \(\mathcal{O}(T^{-1/6})\) with \(\gamma = \mathcal{O}(1/\sqrt{T})\) in large-\(T\) non-convex settings, making \(\boldsymbol{m}_t\) an effective approximation. Our synergistic gradient adjustment strategy ensures that updates to $\boldsymbol{s_w}$ navigate towards an overlapping optimal area of two unrelated objectives.  Using Eqs. \ref{eqn:proj_a} and \ref{eqn:proj_e}, adversarial identity updates are {effectively} redirected into emotional updates and vice versa, ensuring that the output $\boldsymbol{x}_p$ appears natural while incorporating robust target identity's attributes. We provide an illustration in Fig. \ref{fig:grad_proj}.

\textbf{\textit{Training Convergence Theorem}.}  \textit{Suppose that the loss functions defined in Eqs. \ref{eqn:adv} and   \ref{eqn:emoji} satisfy the following {two} conditions: (1) their gradients $\nabla \mathcal{L}_{\text{A/E}}(\boldsymbol{w})$ are bounded; (2) the stochastic gradients are L-Lipschitz. Suppose $\boldsymbol{s_w}$ is updated using SGD with adjusted gradients in Eqs. \ref{eqn:proj_a} and \ref{eqn:proj_e}. Let the learning rate $\eta_t$ be $\frac{\eta_0}{\sqrt{t}}$, we have:}

$\frac{1}{T}\Sigma^{T}_{t=1}\mathbb{E}_{\boldsymbol{x}_o\sim\mathcal{D}}\left[ ||\nabla \mathcal{L}_{\text{A/E}}(\boldsymbol{w}^t) ||_2\right ] \leq \mathcal{O}(\frac{\log T}{\sqrt{T}})$.

For non-convex stochastic optimization, the above theorem shows that training $\boldsymbol{s_w}$ with projected gradients in Eqs. \ref{eqn:proj_a} and   \ref{eqn:proj_e} have a convergence rate $\mathcal{O}(\log T / \sqrt{T})$, but enjoys better performance compared to naively training.

\begin{figure}[t]
\centering
\includegraphics[width=0.38\textwidth]{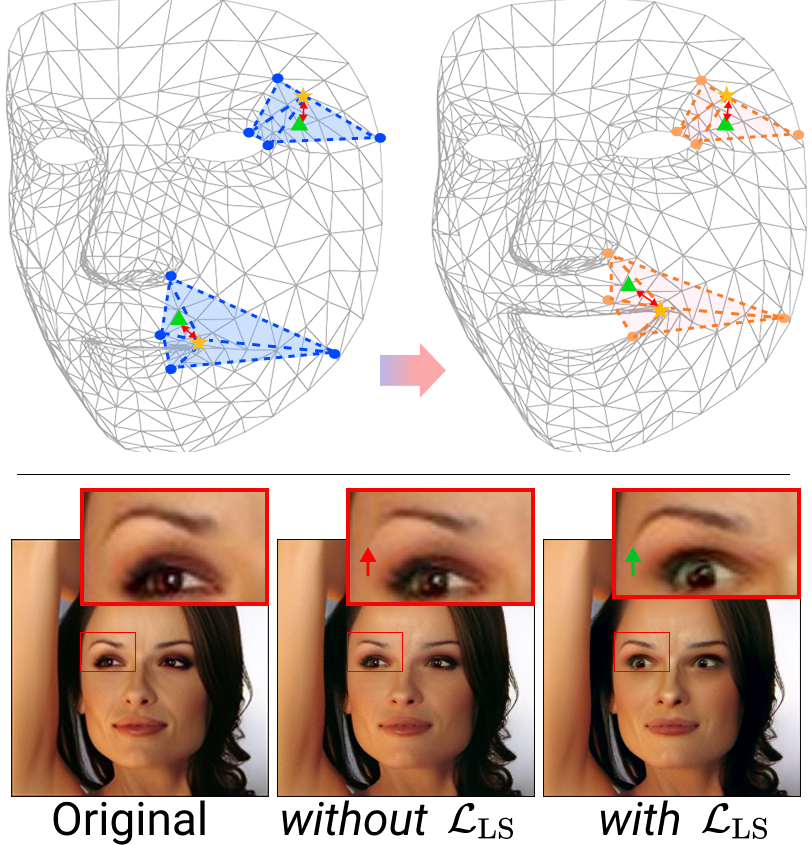}
\caption{\textbf{Top: } Illustration of Laplacian Smoothness Regularization for landmarks (\textcolor{orange}{\Pisymbol{pzd}{72}}) 54$^{th}$ and 26$^{th}$. \textcolor{green}{$\blacktriangle$} represents the average of neighbors (\textcolor{gray}{$\Large \bullet$}) of \textcolor{orange}{\Pisymbol{pzd}{72}}. \textbf{Bottom:} Effects of Laplacian Smoothness; without it, the eyebrow is eroded compared to the original image.
}
\label{fig:laplacian_smooth}
\end{figure}
\begin{figure*}[th!]
\centering
\includegraphics[width=0.78\textwidth]{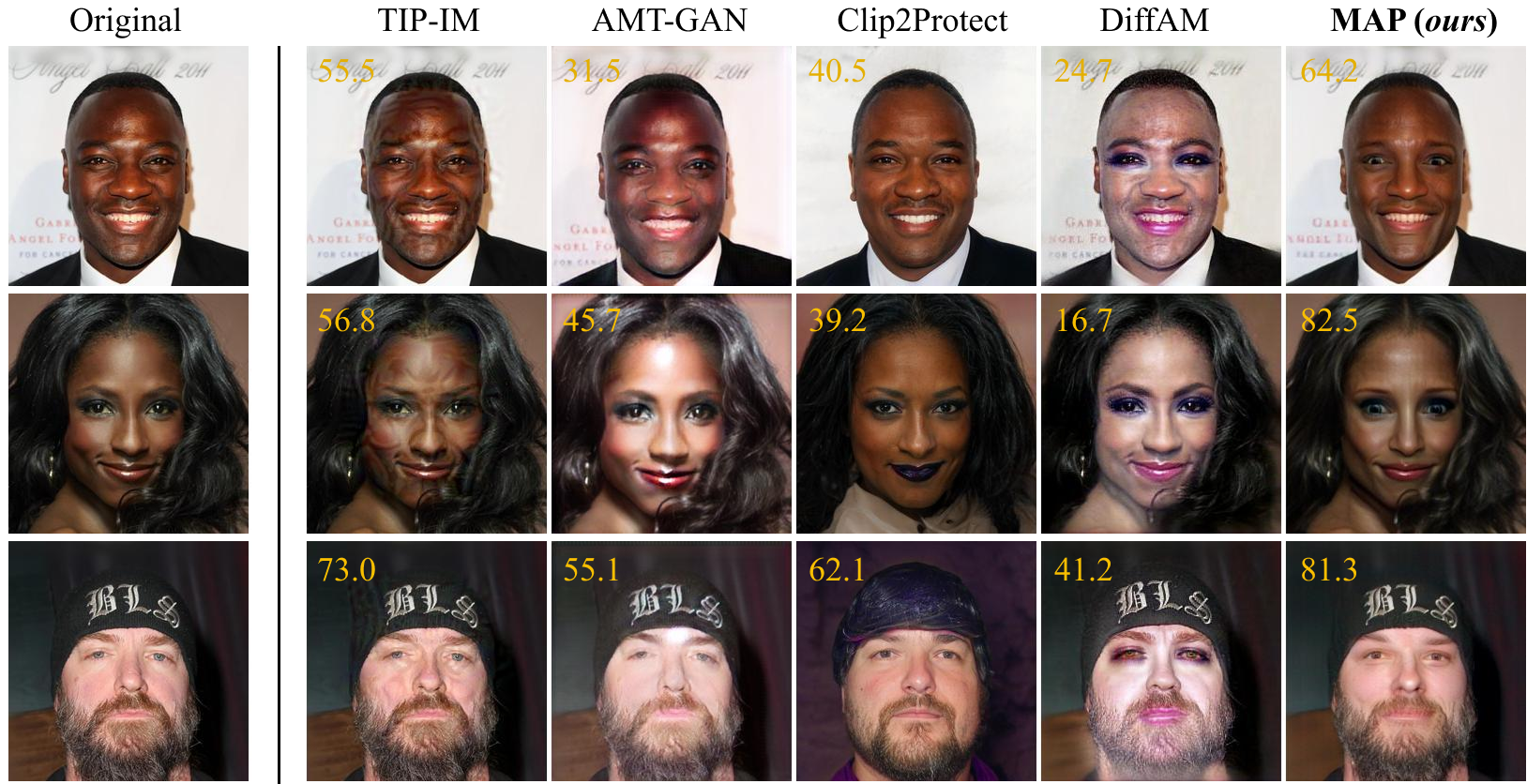}
\caption{{Visualizations of protected face images generated by different facial privacy protection methods on CelebA-HQ}. The {yellow numbers} in each image represent confidence scores returned by Face++. Unlike makeup-based approaches, which may not be suitable for all demographics, our method successfully protects images against malicious FR systems through emotion editing (top to bottom: surprise, surprise, happy), while preserving original details like color grading and background. 
}
\label{fig:baselines_comparison}
\end{figure*}

\textbf{Visual Perception Objective}. To enhance the quality of $\boldsymbol{x}_p$, we adopt a methodology aligned with prior works \cite{diffam, clip2protect}, utilizing LPIPS and $\ell_1$ loss ($\mathcal{L}_\text{LPIPS}$ and $\mathcal{L}_1$).  Distinct from methods that rely on makeup transfer, our method allows for the adjustment of facial attributes through action units, thereby highlighting the shortcomings of pixel-wise perceptual losses. Furthermore, as shown in Fig. \ref{fig:laplacian_smooth},  relying solely on traditional perceptual losses leads to undesirable alterations in the original facial features, such as the length of the eyebrow, resulting in visually unappealing results. To address this, we introduce a Laplacian smoothness loss to ensure that any facial deformations are natural and maintain the integrity of facial landmarks' relative positions,
\begin{align}
    \mathcal{L}_{\text{LS}} (\boldsymbol{w}) =\mathbb{E}_{\substack{\boldsymbol{x}_o\sim \mathcal{D} \\  i\sim \mathcal{K} }}|| \Delta v_o^i- \Delta v_p^i||, \Delta v^i = v^i-\mathbb{E}_{ \mathcal{N}_{i}}v^j,
    \label{eqn:laplacian}
\end{align}
where $v_{o/p}^i$ represent the facial landmarks of the original and protected images, identified by a pretrained landmark regression model.   $\mathcal{N}_i$ denotes the neighbors of vertex $v^i$  formed by the Delaunay triangulation algorithm \cite{delaunay}, and $\mathcal{K}$ is a set of selected landmarks for applying the smoothness loss. An illustration of  $\mathcal{L}_{\text{LS}} $ is shown in Fig. \ref{fig:laplacian_smooth} - Top. In our experiments, we also observe that applying Laplacian smoothness makes our training more stable, and helps avoid mode collapse compared to not using it.

\textbf{Score matching}. While the score network $\boldsymbol{s_w}$ is fine-tuned by objectives in Eqs. \ref{eqn:adv}, \ref{eqn:emoji}, and  \ref{eqn:laplacian}, so far, those objectives are applied to data space (timestep of $0$) of $\boldsymbol{s_w}$. Meanwhile, study by \citet{guo2024gradient} has shown that a pretrained score network serves as a prior regularization during generation, which prompts us to fine-tune the score network using the facial training data by minimizing score matching loss in its latent space as follows:
\begin{equation}
    \mathcal{L}_D(\boldsymbol{w}) = \sum_{t=1}^{\tau}\mathbb{E}_{\boldsymbol{x}_o\sim \mathcal{D}, \epsilon_t} \left [ ||\epsilon_t -\boldsymbol{s_w}(\boldsymbol{x}^{t}_o, t)|| \right],
\end{equation}
where $\epsilon_t$ is the noise added to $\boldsymbol{x}_o$ to produce $\boldsymbol{x}^t_o$.  During training, we alternately optimize $\mathcal{L}_D$ with other objectives.
\begin{table*}[th!]
\centering
\resizebox{0.84\textwidth}{!}{%
\begin{tabular}{ll|cccc|cccc|c}
\hline
\multicolumn{2}{c|}{\multirow{2}{*}{Method}}  & \multicolumn{4}{c|}{{CelebA-HQ}} & \multicolumn{4}{c|}{{LADN-Dataset}} & \multirow{2}{*}{Avg.} \\ \cline{3-10}
& & {IRSE50} & {IR152} & {FaceNet} & {MobileFace} & {IRSE50} & {IR152} & {FaceNet} & {MobileFace} &  \\ \hline \hline
 \multicolumn{1}{c|}{} & Clean & 7.29 & 3.80 & 1.08 & 12.68 & 2.71 & 3.61 & 0.60 & 5.11 & 4.61 \\ \hline
 \multicolumn{1}{c|}{\multirow{4}{*}{Noise-based}} & PGD 
 & 36.87 & 20.68 & 1.85 & 43.99 & 40.09 & 19.59 & 3.82 & 41.09 & 25.60 \\ 
 \multicolumn{1}{c|}{} & MI-FGSM 
 & 45.79 & 25.03 & 2.58 & 45.85 & 48.90 & 25.57 & 6.31 & 45.01 & 30.63 \\  
\multicolumn{1}{c|}{} & TI-DIM 
& 63.63 & 36.17 & 15.30 & 57.12 & 56.36 & 34.18 & 22.11 & 48.30 & 41.64 \\
\multicolumn{1}{c|}{} & TIP-IM 
& 54.40 & 37.23 & 40.74 & 48.72 & 65.89 & 43.57 & 63.50 & 46.48 & 50.06 \\ \hline
 \multicolumn{1}{c|}{\multirow{4}{*}{Makeup-based}} & Adv-Makeup 
 & 21.95 & 9.48 & 1.37 & 22.00 & 29.64 & 10.03 & 0.97 & 22.38 & 14.72 \\  
\multicolumn{1}{c|}{} & AMT-GAN 
& 76.96 & 35.13 & 16.62 & 50.71 & 89.64 & 49.12 & 32.13 & 72.43 & 52.84 \\  
\multicolumn{1}{c|}{} & CLIP2Protect 
& 81.10 & 48.42 & 41.72 & 75.26 & 91.57 & 53.31 & 47.91 & 79.94 & 64.90 \\  
\multicolumn{1}{c|}{} & DiffAM 
& 92.00 & 63.13 & 64.67 & 83.35 & 95.66 &66.75 & 65.44 & 92.04 & 77.88 \\ \hline
  \rowcolor{backcolour}\multicolumn{1}{c|}{Emotion-based} & \SystemName~  (\textit{\textbf{ours}})  & \textbf{93.30}& \textbf{78.98} & \textbf{72.35} & \textbf{92.50} & \textbf{96.65} &\textbf{91.16}& \textbf{86.43}  &\textbf{96.49}  & \textbf{88.48} 
\\
\bottomrule
\bottomrule
\end{tabular}
}
\caption{{Protect Success Rate (PSR) with FAR@1e-2 for black-box setting on CelebA-HQ and LADN dataset.} 
}
\label{tab:asr_results}
\end{table*}

\begin{table}[ht]
\centering
\resizebox{0.96\columnwidth}{!}{%
\begin{tabular}{l| l@{\hspace{3pt}}r |l@{\hspace{3pt}}r| l@{\hspace{3pt}}r |l@{\hspace{3pt}}r}
\toprule
Method & \multicolumn{2}{c|}{IRSE50} & \multicolumn{2}{c|}{IR152} & \multicolumn{2}{c|}{FaceNet} & \multicolumn{2}{c}{MobileFace} \\ 
\cmidrule(r){2-3} \cmidrule(l){4-5} \cmidrule(lr){6-7} \cmidrule(l){8-9} 
       & R1-T & R5-T & R1-T & R5-T & R1-T & R5-T & R1-T & R5-T \\
\midrule
\midrule
TIP-IM        & 16.2 & 51.4 & 21.2 & 56.0 &  8.1 & 35.8 &  9.6 & 24.0 \\
CLIP2Protect & 24.5 & 64.7 & 24.2 & 65.2 & 12.5 & 38.7 & 11.8 & 28.2 \\
SD4Privacy & 15.6 & 26.8 & 23.4 & 41.2 &33.6 & 53.8 & 31.8 & 49.8 \\
GIFT  & 21.2 & 57.2 & 34.6  & 49.4 &  33.2 & 65.6 & 41.2 & 67.6 \\
Adv-CPG  & 24.4 & 56.4 & 33.8 & 51.2 & 36.6 &  67.4 &  43.4 & 70.4 \\
\hline

\rowcolor{backcolour}\SystemName~ (\textit{\textbf{ours}})   & \textbf{57.9}  & \textbf{83.2} & \textbf{54.0} & \textbf{61.8} & \textbf{45.0 }& \textbf{70.7} & \textbf{50.3} & \textbf{74.5} \\
\bottomrule
\bottomrule
\end{tabular}%
}
\caption{{Protection success rate (PSR) of impersonation attacks under the face identification task on  CelebA-HQ. }}
\label{tab:identification_rates}
\end{table}

\textbf{End-to-end training procedure.} Our end-to-end training objective is provided as follows: 
\begin{align}
     \mathcal{L} = {\gamma_a\mathcal{L}_\text{A}  + \gamma_e\mathcal{L}_\text{E}} 
     + {\gamma_{\text{lpips}}\mathcal{L}_\text{LPIPS} + {\gamma_{1}\mathcal{L}_\text{1}} +  \gamma_{\text{ls}}\mathcal{L}_\text{LS} } +  {\gamma_{d}\mathcal{L}_\text{D}},
    \label{eqn:overall_loss}
\end{align}
 \\
where $\gamma_i$'s are hyperparameters that balance the contributions of each objective. During training, the adversarial emotion term is optimized using projected gradients, and the score matching is alternately optimized with other terms. 

\section{Experiments}
\label{sec:exp}

\subsection{Experimental Settings}
\label{subsec:exp_setting}
\textbf{Datasets}. We evaluate our approach using face verification and identification tests, alongside impersonation attack scenarios \cite{diffam}. For face verification, we employ the CelebA-HQ \cite{celeba2016} and LADN \cite{gu2019ladn} datasets for impersonation attacks. Following the standard protocol by \citet{amt_gan}, in CelebA-HQ, we use 1,000 images to impersonate 4 distinct  target identities. In LADN, we divide the available 332 images into 4 groups, each group targeting different identities for impersonation. 

\textbf{Target model}. Consistent with prior studies \cite{diffam}, we assess facial privacy protection by attacking four FR models with diverse backbones under black-box settings. These models include IRSE50, IR152, FaceNet, and MobileFace. In each experiment, three models serve as surrogates for training, and one for black-box evaluation.

\textbf{Evaluation metric}. To assess privacy protection effectiveness, we employ the Protection Success Rate (PSR), defined as the fraction of protected faces misclassified by the black-box FR system. We set the threshold at the False Acceptance Rate (FAR) of 0.01 for each model. Additionally, we evaluate the image quality of protected faces using standard metrics, including FID \cite{heusel2017fidscore}, PSNR (dB), and SSIM \cite{wang2004ssim}. 

\textbf{Implementation Details}.  We fine-tune our score network on 200 images from each dataset, using AdamW  as the optimizer with a learning rate of 4e-6 and 5e-7 for CelebA and LADN, respectively. The training is conducted over 10 epochs with a batch size of 4 on 4 NVIDIA RTX 3090 24GB GPUs. To mask original identities, we employ the ``surprised" expression, which is context-dependent and can convey either positive or negative emotions. We set  $\{ \gamma_\text{lpips}=0.1, \gamma_1=0.5, \gamma_\text{ls}=4, \gamma_{d}=0.05\}$ for CelebA-HQ and double them for LADN; also  $\{\lambda=0.95, \gamma_a=0.5, \gamma_e=0.08\}$ is commonly used for both datasets. 

\subsection{Experimental Results}
\label{subsec:results}
\textbf{Face Verification Task}. Our quantitative results in terms of Protection Success Rate (PSR) for impersonation attacks under face verification tasks are presented in Table \ref{tab:asr_results}. As shown, \SystemName\  demonstrates superior performance across black-box models on both the CelebA-HQ and LADN datasets. Compared to noise-based and makeup-based approaches, it significantly improves performance by 38\% and 11\%, respectively, on average.

\textbf{Face Identification Task}. We select 500 subjects from CelebA-HQ, each represented by a pair of images, and additionally integrate four target identities into the gallery set. For face identification, we measure the Rank-N targeted identity success rate, which determines whether the target image $\boldsymbol{x}_t$ appears at least once among the top N candidates in the gallery.  We present PSR at both Rank-1 and Rank-5 settings in Table \ref{tab:identification_rates}. Our approach consistently outperforms recent SoTA baselines in both settings, with average performance improvements of 33\% and 23\%, respectively.

\textbf{Quantitative Comparison}. We report the performance of various methods in terms of perception quality in Table \ref{tab:quant_results}. The results are averaged across both CelebA-HQ and LADN datasets on the verification task. While Adv-Makeup \cite{adv_makeup} update, which only synthesizes eyeshadow to obscure the source identity, reveals the highest scores in all quantitative measures, it has minimal PSR. Our method exhibits the lowest  FID scores compared to all other makeup-based baselines and achieves the highest PSR gain.

\begin{figure}[t!]
\centering
\includegraphics[width=0.40\textwidth]{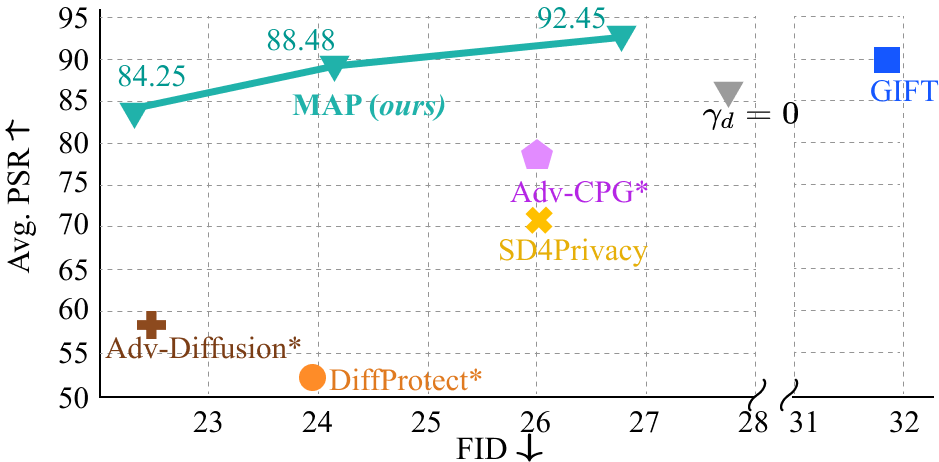}
\caption{Trade-off between PSR and FID. \SystemName{} achieves a better trade-off and outperforms five freeform baselines: DiffProtect, SD4Privacy, Adv-Diffusion, GIFT, and Adv-CPG.}
\label{fig:psr_vs_fid}
\end{figure}

\textbf{Comparison with Freeform Approaches.} We explore the trade-off between quantitative (PSR under face verification task) and qualitative performance  by simply adjusting the values of $\gamma_d$. Additionally, we include the performance of five freeform attribute baselines, DiffProtect \cite{diffprotect}, SD4Privacy \cite{an2024sd4privacy}, Adv-Diffusion \cite{liu2024adv}, GIFT \cite{gift_mm}, and Adv-CPG~\cite{wang2025adv}. As shown in Fig. \ref{fig:psr_vs_fid}, \SystemName\ provides a higher PSR at similar FID scores compared to all freeform baselines. Meanwhile, Fig. \ref{fig:psr_vs_fid} also shows that with $\gamma_d = 0$, our method has similar quantitative performance to GIFT but cannot achieve its best performance, indicating the crucial role of optimizing the score matching. 
\begin{table}[t!]
\centering
\resizebox{.40\textwidth}{!}{%
\begin{tabular}{l|ccc|c}
\hline
{Method} & {FID $\downarrow$} & {PSNR  $\uparrow$ } & {SSIM  $\uparrow$ } & {PSR Gain  $\uparrow$ }
\\ \hline\hline
Adv-Makeup 
& 4.22 & 34.51 & 0.985 & 0 \\ 
\hline
AMT-GAN  
& 34.44 & 19.50 & 0.787 & 38.12\\ 
CLIP2Protect 
& 26.62 & 19.31 & 0.750 & 50.18\\ 
 DiffAM 
 & 26.10 & 20.52 & {0.886} & 63.13\\ \hline
   \rowcolor{backcolour}\SystemName~ (\textit{\textbf{ours}}) &{24.21} &{29.07} & 0.876 &{73.76} \\
\bottomrule
\bottomrule
\end{tabular}
}
\caption{{Quantitative evaluations of image quality}. PSR Gain is absolute gain in PSR relative to Adv-Makeup.}
\label{tab:quant_results}
\end{table}
\begin{figure}[t!]
\centering
\includegraphics[width=0.46\textwidth]{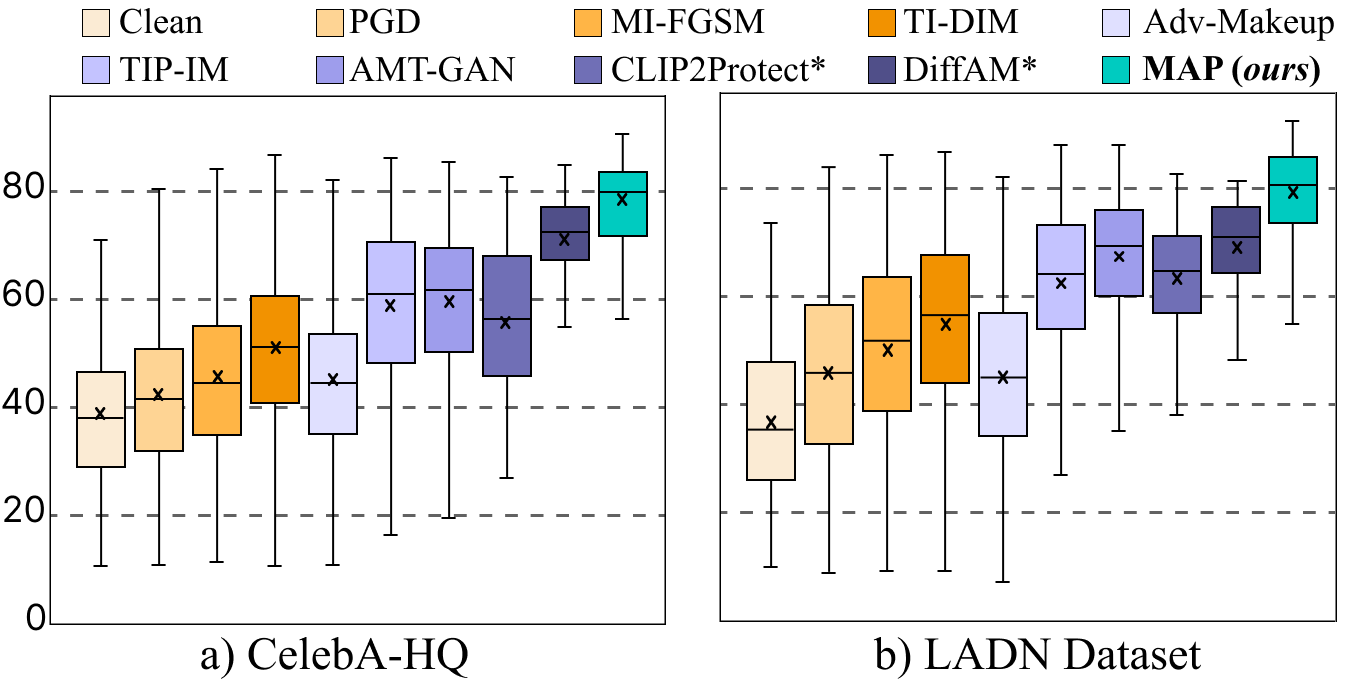}
\caption{ {Average confidence score from the real-world face verification API, Face++, for impersonation attacks}. 
}

\label{fig:oneline_eval}
\end{figure}

\textbf{Real-world Effectiveness}.  To  validate the effectiveness of \SystemName\  against unauthorized FR systems, we conducted experiments using a commercial API, namely Face++, with a face verification test. The API returns confidence scores between two images on a scale from 0 to 100, where a higher score indicates greater similarity. We run the experiment with 1,000 images from the CelebA dataset and 324 images from the LADN dataset. As shown in Fig. \ref{fig:oneline_eval}, our proposed method achieved the highest confidence scores, with improvements over the second-best method of 8\% and 10\% on the CelebA-HQ and LADN datasets, respectively.

\label{subsec:ablations}

\begin{table}[t!]
\centering
\resizebox{.34\textwidth}{!}{%
\begin{tabular}{l|c|ccc}
\hline
\multirow{2}{*}{{Settings}} & \multirow{2}{*}{{FID}$\downarrow$ } & \multicolumn{3}{c}{{PSR$\uparrow$  with FAR}}  \\
\cline{3-5}
& & {@1e-1}  & {@1e-2}  & {@1e-3} \\ \hline\hline
w/o emotion  &24.1&  95.6 & 85.0 & 67.0 \\ 
\hline
w/o grads proj.& 24.5 &  90.5  & 71.2 & 50.9\\ 
w/o EMA &22.5& 93.6  & 76.5 & 56.8  \\  
w/o $\mathcal{L}_\text{LS}$ & 23.6&  94.8  & 78.8 & 56.4 \\ \hline
\rowcolor{backcolour}\SystemName~ (\textit{{ours}})  &21.5& 95.4  & 80.3 & 60.5 \\ \hline\hline
\end{tabular}%
}
\caption{Ablation studies on different optimization settings.}
\label{tab:settings_comparison}
\end{table}

\textbf{Impact of Loss Components}.  We evaluate on the largest black-box model, IR152 \cite{deng2019ir152}, using the first identity from the CelebA-HQ dataset to validate the effect of each proposed loss component. As shown in Table \ref{tab:settings_comparison}, without our proposed gradient projection, the model exhibits suboptimal performance, illustrating the negative transfer effects during optimization.   Meanwhile, obfuscating the target identity without the use of makeup or emotional transformations can obtain higher PSR yet results in poorer qualitative scores due to erosion of facial attributes.

\begin{figure}[t!]
\centering
\includegraphics[width=0.38\textwidth]{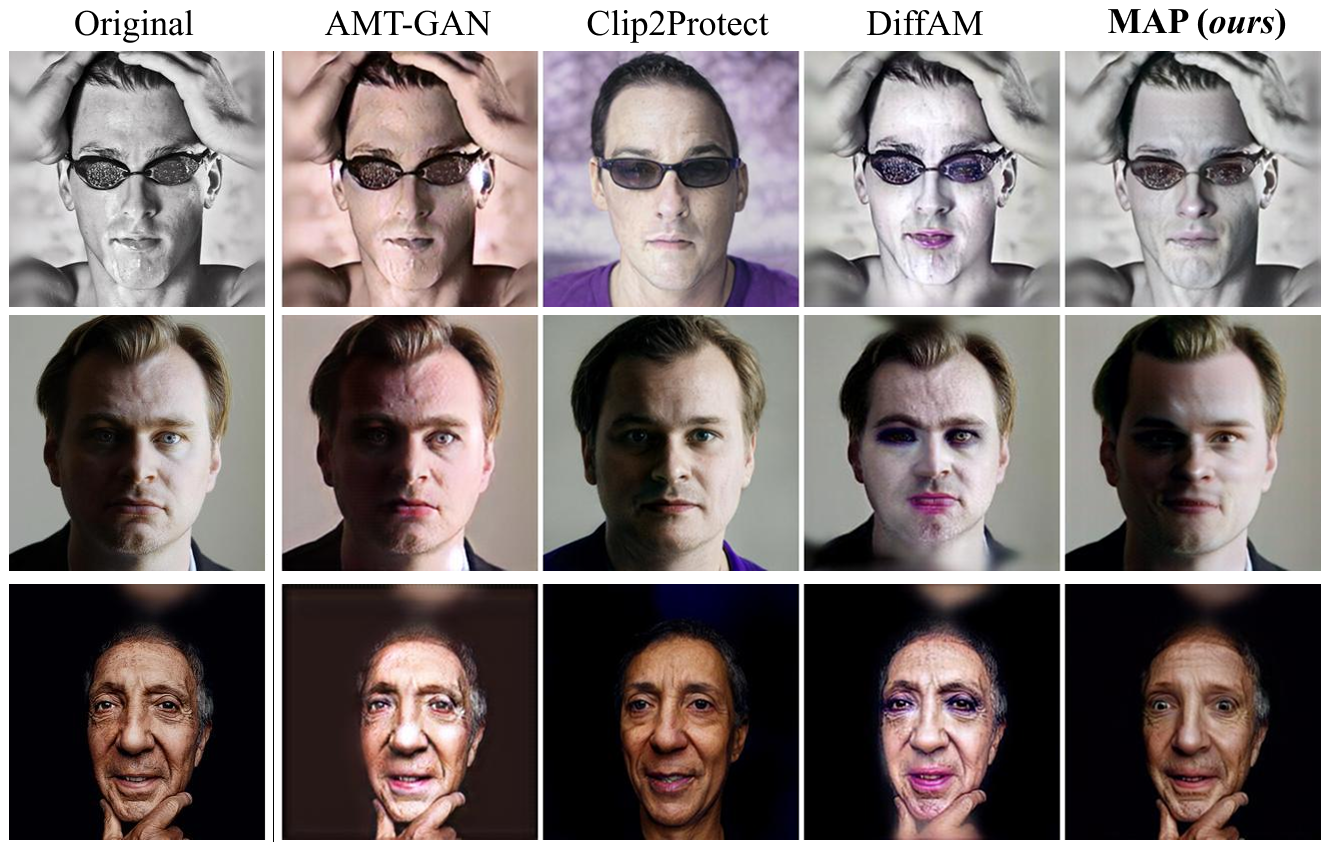}
\caption{{Illustration of uncommon photography styles: monochrome, Rembrandt lighting, and backlighting}.
}
\label{fig:uncommon_lot}
\end{figure}

\textbf{Uncommon Portrait Photography Styles.} We compare our \SystemName{} method against baselines across uncommon photographic styles, namely monochrome, Rembrandt lighting, and backlighting, as illustrated in Fig. \ref{fig:uncommon_lot}. As shown, unlike makeup-based methods, our approach exhibits minimal impact on monochrome images while preserving fine details, color grading, and object naturalness in protected images across all tested styles. This demonstrates the robustness and versatility of our method across diverse scenarios.

\begin{figure}[t!]
\centering
\includegraphics[width=0.44\textwidth]{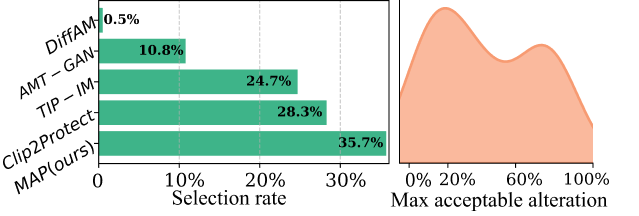}
\caption{\footnotesize{Our human evaluation from 25 participants.}}
\label{fig:survey}

\end{figure}

\textbf{Human Study}. To evaluate practical applicability, we conducted a user study to compare users’ preferences for our method against previous baselines. We asked 25 participants to answer 25 questions about their preferred editing method for use on social networks while all method names were replaced with dummy text. Our method was favored in 35.7\% of the cases (Fig.\ref{fig:survey}-left). Additionally, most users preferred alterations of 20\% or less (Fig.\ref{fig:survey}-right), highlighting the suitability of \SystemName’s subtle modifications compared to  makeup, hairstyle, or facial attribute eddits.
\section{Conclusion}
\label{sec:conclusion}
    This paper presents a new approach to disguise the original identity as a target identity in arbitrary portrait images, enhancing privacy protection for facial images on online platforms. Departing from prior methods, we transform expressions in the source image to redirect recognition from the original identity to a target one, thwarting adversarial FR systems. To reconcile conflicts between emotion and identity learning, we project their mini-batch gradients onto mutual empirical gradients, ensuring cohesive optimization. We also introduce a perceptual objective, Laplacian smoothness, combined with score matching loss to maintain image naturalness. Experiments show our method surpasses noise-based, makeup-based, and freeform attribute approaches in quantitative metrics and qualitative fidelity, underscoring its potential as a robust privacy safeguard. 
\section{Acknowledgments}
This work was partly supported by Institute for Information \& communication Technology Planning \& evaluation (IITP) grants funded by the Korean government MSIT: (RS-2022-II221199, RS-2022-II220688, RS-2019-II190421, RS-2023-00230337, RS-2024-00437849, RS-2021-II212068, and RS-2025-02263841). Also, this work was supported by the Cyber Investigation Support Technology Development Program (No.RS-2025-02304983) of the Korea Institute of Police Technology (KIPoT), funded by the Korean National Police Agency. Lastly, this work was supported by the National Research Foundation of Korea (NRF) grant funded by the Korea government (MSIT) (No.RS-2024-00356293).
\bibliography{aaai2026}

\end{document}